\documentclass[10pt,twocolumn,letterpaper]{article}
\usepackage{multirow}
\usepackage{stfloats}
\usepackage[accsupp]{axessibility}
\usepackage{cvpr}              

\definecolor{cvprblue}{rgb}{0.21,0.49,0.74}
\usepackage[pagebackref,breaklinks,colorlinks,allcolors=cvprblue]{hyperref}

\def\paperID{10452} 
\def\confName{CVPR}
\def\confYear{2025}

\title{HuMoCon: Concept Discovery for Human Motion Understanding
}

\author{
Qihang Fang$^{1,2}$ \quad Chengcheng Tang$^{2}$ \quad Bugra Tekin$^{2}$ \quad Shugao Ma$^{2}$ \quad Yanchao Yang$^{1}$ \\
$^1$The University of Hong Kong \quad $^2$Meta \\
{\tt\small qihfang@gmail.com, \{chengcheng.tang,bugratekin,shugao\}@meta.com, yanchaoy@hku.hk}
}

\begin{document}


\twocolumn[{%
\renewcommand\twocolumn[1][]{#1}%
\maketitle

\begin{center}
\vspace{-5mm}
  \centering
  \includegraphics[width=1.0\linewidth]{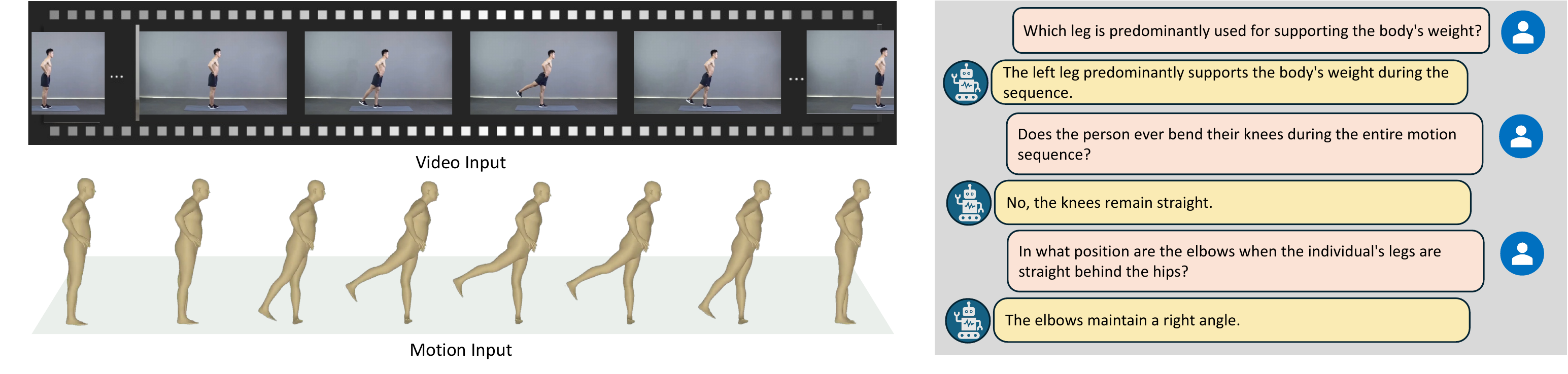}
   \captionof{figure}{We present HuMoCon for concept discovery and human motion understanding with video or motion input. To address the challenge of feature misalignment and high-frequency information loss, we propose a novel feature alignment strategy and an advanced masked auto-encoder reconstructing velocity. HuMoCon empowers effective motion concept discovery and accurate Question Answering tasks, significantly outperforming state-of-the-art methods qualitatively and quantitatively.}
   \label{fig:teaser}
\end{center} %
}]

\begin{abstract}

We present HuMoCon, a novel motion-video understanding framework designed for advanced human behavior analysis. The core of our method is a human motion concept discovery framework that efficiently trains multi-modal encoders to extract semantically meaningful and generalizable features. HuMoCon addresses key challenges in motion concept discovery for understanding and reasoning, including the lack of explicit multi-modality feature alignment and the loss of high-frequency information in masked autoencoding frameworks. Our approach integrates a feature alignment strategy that leverages video for contextual understanding and motion for fine-grained interaction modeling, further with a velocity reconstruction mechanism to enhance high-frequency feature expression and mitigate temporal over-smoothing. Comprehensive experiments on standard benchmarks demonstrate that HuMoCon enables effective motion concept discovery and significantly outperforms state-of-the-art methods in training large models for human motion understanding. We will open-source the associated code with our paper.

\end{abstract}    
\section{Introduction}

Understanding human behavior 
is crucial for gaining deeper insights into human actions and reasoning, 
which are foundational for developing human-centered AI systems \cite{chen2024motionllm,han2024autoad,li2024llava,starke2022deepphase} 
that effectively address real-world tasks \cite{wang2023learning,fang2022towards,xiao2023unified}. 
Recent advancements in large language models (LLMs) 
have led to significant progress in video- and motion-based methods \cite{lin2023video,chen2024motionllm,fang2024cigtime,li2024llava,xu2024pllava} 
for analyzing human behavior. 
Despite these advancements, 
achieving a precise and fine-grained understanding of human actions remains a persistent and complex challenge.

Human behavior is often represented as motion sequences 
structured based on the kinematics of the human skeleton \cite{loper2023smpl,jiang2024smplx}. 
Traditional approaches to analyzing motion sequences 
have predominantly focused on classification \cite{liu2016spatio,du2015hierarchical,liu2017enhanced,ke2017new,yan2018spatial,shi2019two,guo2022contrastive,lin2023actionlet}, 
relying on predefined action categories. 
While effective for specific tasks, 
these methods often lack the flexibility for deeper or more nuanced analysis. 
Other methods employ statistical models to assess motion quality \cite{parmar2019and,bai2022action,liu2021towards,chen2020pose,fieraru2021aifit}, 
but they require significant expert knowledge, limiting scalability. 

Recent advancements in language models \cite{lin2023video,xu2024pllava,munasinghe2023pg} 
have opened new possibilities in motion analysis, 
enabling tasks such as motion generation and motion captioning based on free-text inputs \cite{starke2019neural,cervantes2022implicit,guo2020action2motion,petrovich2021action,starke2022deepphase}, 
advancing the field of motion analysis. 
However, obtaining high-quality motion sequences and annotations remains costly, and motion sequences generally lack information about the surrounding environment, limiting models' ability to understand human-environment interactions. 

In contrast, 
video data offers rich visual context and is easier to collect. It has been widely used for action classification \cite{taylor2010convolutional,tran2015learning,feichtenhofer2019slowfast} and video Q\&A \cite{lin2023video,li2024llava,chen2024motionllm}. However, video data often contains irrelevant elements that can hinder accurate content analysis. 
For example, a model might learn from training data that most running occurs in stadiums, 
leading it to mistakenly classify other stadium activities as running during testing.

In recent years, 
significant progress has been made in developing multi-modality models \cite{zhu2023languagebind,chen2024motionllm}. 
Integrating multiple modalities, 
as opposed to relying on a single modality, 
enhances the network’s ability to learn fine-grained and semantic features. 
For instance, LART \cite{rajasegaran2023benefits} combines video and motion data to ensure that video-based action classification focuses on the human body rather than overemphasizing environmental context. 
MotionLLM \cite{chen2024motionllm} expands large language models (LLMs) to video and motion, 
building a model that can understand both. 
MotionLLM demonstrated that learning from video and motion improves the network's ability to analyze human behavior. 
However, MotionLLM implicitly aligns video and motion using motion-video pairs during LLM training without optimizing the encoder for explicit cross-modal alignment. 
To address this limitation and further enhance feature representation, 
we propose explicitly aligning features 
from different modalities 
during the encoding process 
within a human motion concept discovery framework.

Achieving effective cross-modal alignment and feature representation 
relies heavily on robust encoder training methods. 
Many existing studies~\cite{wang2022internvideo,wang2024internvideo2,zhu2023languagebind} adopt a masked autoencoder framework~\cite{he2022masked} 
when training encoders. 
This approach involves applying a mask to a certain proportion of the input, 
extracting features from the masked data using the encoder, 
and reconstructing the original unmasked input with a decoder. 
While effective in learning high-quality features, 
this masking strategy inherently causes the loss of high-frequency information, 
resulting in temporally over-smoothed reconstructions that hinder the LLM's ability to understand human behavior.~\cite{lucas2022posegpt}
To address this issue, 
we propose reconstructing velocity as a means to preserve high-frequency information. 
Following InfoCon~\cite{liu2024infocon}, 
we design a VQ-VAE-based loss function that helps retain the fine-grained features of the original masked autoencoder while further enhancing high-frequency feature expression. 
This approach significantly boosts the subsequent LLM's capacity to model and understand human motion effectively.

In summary, 
we present HuMoCon, 
an effective human motion concept discovery pertaining framework 
for advanced human behavior understanding from both video and motion data. 
Our key contributions include: 
(i) a novel feature alignment strategy that 
explicitly leverages video data for contextual understanding and 
motion data for fine-grained interaction modeling, 
and (ii) an enhanced masked autoencoder framework 
incorporating velocity reconstruction to significantly improve high-frequency feature expression and mitigate temporal over-smoothing. 
To evaluate the effectiveness of our proposed method, 
we conduct experiments on the Activity-QA~\cite{yu2019activityqa} and BABEL-QA~\cite{endo2023motion} benchmarks~\cite{punnakkal2021babel}, 
comparing our approach with the state-of-the-art methods. 
Comprehensive quantitative and qualitative evaluations 
demonstrate that HuMoCon achieves state-of-the-art 
performance on standard benchmarks for motion-video understanding.

\section{Related Work}

\subsection{Human Motion Understanding}

The core of human motion understanding 
is to learn the semantics of motion sequences or videos. 
In the early stages, 
research in this field is
often based on predefined action categories. 
Such work can be divided into two types: 
action classification \cite{liu2016spatio,du2015hierarchical,liu2017enhanced,ke2017new,yan2018spatial,shi2019two,guo2022contrastive,lin2023actionlet} 
and motion generation targeted at specific categories \cite{jiang2021hand,kritsis2022danceconv,holden2017phase,starke2020local,yang2022learning,xu2024dexterous,starke2019neural,cervantes2022implicit,guo2020action2motion,petrovich2021action,starke2022deepphase}. 
For action recognition, 
several works \cite{liu2016spatio,du2015hierarchical} 
utilize Recurrent Neural Networks (RNN) to extract temporal information from human joint sequences, 
beyond the CNN-based ones \cite{liu2017enhanced,ke2017new}. 
Subsequent works~\cite{yan2018spatial,shi2019two,guo2022contrastive,lin2023actionlet}  propose representing motion data as a graph structure, 
enabling the use of spatio-temporal graph convolutional networks (ST-GCNs) to effectively capture the connectivity and interactions between human joints. 
For motion generation, several studies~\cite{jiang2021hand,kritsis2022danceconv,holden2017phase,starke2020local,yang2022learning,xu2024dexterous} have concentrated on generating specific, single actions. In contrast, other works~\cite{starke2019neural,cervantes2022implicit,guo2020action2motion,petrovich2021action,starke2022deepphase} aim to generate a broader range of diverse action types, addressing the challenge of variability in motion synthesis. 

In recent years, 
more research has shifted towards captioning ~\cite{jiang2023motiongpt,guo2022tm2t}, generation ~\cite{jiang2023motiongpt,guo2022tm2t,petrovich2022temos,athanasiou2022teach,guo2022generating,zhang2024motiondiffuse,tevet2022human,kim2023flame,zhang2023t2m,zhou2024avatargpt}, 
and analysis~\cite{delmas2023posefix,fang2024cigtime,chen2024motionllm,feng2024chatpose} based on free-form text. 
Leveraging alignment of motion latent features with the CLIP~\cite{radford2021learning} embedding space, MotionCLIP~\cite{tevet2022motionclip} enables the generation of out-of-distribution motions. 
Transformers~\cite{petrovich2022temos,athanasiou2022teach,guo2022generating,guo2022tm2t,zhang2023t2m} and diffusion~\cite{zhang2024motiondiffuse,tevet2022human,kim2023flame} models are also utilized for text-conditioned motion generation.
Moreover, MotionGPT~\cite{jiang2023motiongpt} and AvatarGPT~\cite{zhou2024avatargpt} utilize large language models (LLMs) to handle multiple motion-related tasks simultaneously. 
Despite these advancements, research focused specifically on motion analysis remains relatively limited.
PoseFix~\cite{delmas2023posefix} and CigTime~\cite{fang2024cigtime} are proposed to annotate the corrective instructions from source to target for pose and motion sequences.
MotionLLM~\cite{chen2024motionllm} proposes a method to analyze both the video and motion sequences with detailed spatial-temporal awareness and reasoning abilities. 
However, MotionLLM relies on implicit alignment through paired data and lacks explicit cross-modal feature alignment. 

\subsection{Video Understanding}

In addition to motion understanding, video understanding has emerged as a prominent research area , attracting extensive attention in recent years. Early works on video classification and segmentation relied on handcrafted features~\cite{dollar2005behavior,liu2016spatio,vemulapalli2014human,wang2013action} or convolutional neural networks~\cite{taylor2010convolutional,tran2015learning,feichtenhofer2019slowfast}. Beyond classification, video analysis~\cite{parmar2019and,bai2022action,liu2021towards,chen2020pose,fieraru2021aifit,han2023autoad,han2024autoad,zhou2024streaming,xie2024autoad,salewski2023zero,lin2023video,xu2024pllava,munasinghe2023pg,li2024llava} has become a highly active field of research.
Action quality assessment has been the focus of several works~\cite{parmar2019and,bai2022action,liu2021towards,chen2020pose,fieraru2021aifit}, with methods such as Aifit~\cite{fieraru2021aifit} and PoseTrainer~\cite{chen2020pose} proposed to correct user actions. These approaches evaluate actions in conjunction with complex textual instructions~\cite{parmar2019and,bai2022action,liu2021towards}. Other studies~\cite{han2023autoad,han2024autoad,zhou2024streaming,xie2024autoad,salewski2023zero} aim to generate descriptive narratives for videos, broadening the scope of video understanding.
Recently, many works~\cite{lin2023video,xu2024pllava,munasinghe2023pg,li2024llava,zhang2023video,bai2023qwen} have leveraged large language models (LLMs) to understand video content and perform complex reasoning tasks. These methods combine video processing with the advanced reasoning capabilities of LLMs, further advancing the state-of-the-art in video analysis.
Video-LLaVA~\cite{lin2023video} build a model to perform visual reasoning capabilities on images and videos simultaneously. Follow-up works~\cite{xu2024pllava,munasinghe2023pg,li2024llava} extend Video-LLaVA for faster inference and more accurate reasoning ability. 
Unlike typical video understanding algorithms, 
which rely heavily on visual context alone, 
our approach integrates motion data directly, 
enhancing the model's ability to understand human-centric actions.



\subsection{Multi-modality Pre-training}

Multi-modality pre-training has primarily focused on aligning vision and language representations. CLIP~\cite{radford2021learning} pioneered this effort by using contrastive learning to align image and text features on large-scale datasets. Subsequent works~\cite{luo2022clip4clip,elizalde2023clap,zhang2022pointclip,tevet2022motionclip} extended CLIP’s framework to other modalities, including video~\cite{luo2022clip4clip}, audio~\cite{elizalde2023clap} and motion~\cite{tevet2022motionclip}. 

Recent studies, such as VALOR~\cite{chen2023valor} and VAST~\cite{chen2023vast}, demonstrated that cross-modal alignment not only enhances model robustness but also maintains high performance across modalities. Building on this, follow-up works~\cite{zhu2023languagebind,girdhar2023imagebind} have aimed to expand the number of aligned modalities, further advancing multi-modality research.

In the context of motion and video alignment, MotionLLMs~\cite{chen2024motionllm} is the first work to address this challenge. However, it relies solely on implicit alignment during the fine-tuning stage of large language models.

\section{Method}
\label{sec:method}

\subsection{Overview}
\begin{figure*}[!t]
  \centering
  \includegraphics[width=.87\linewidth]{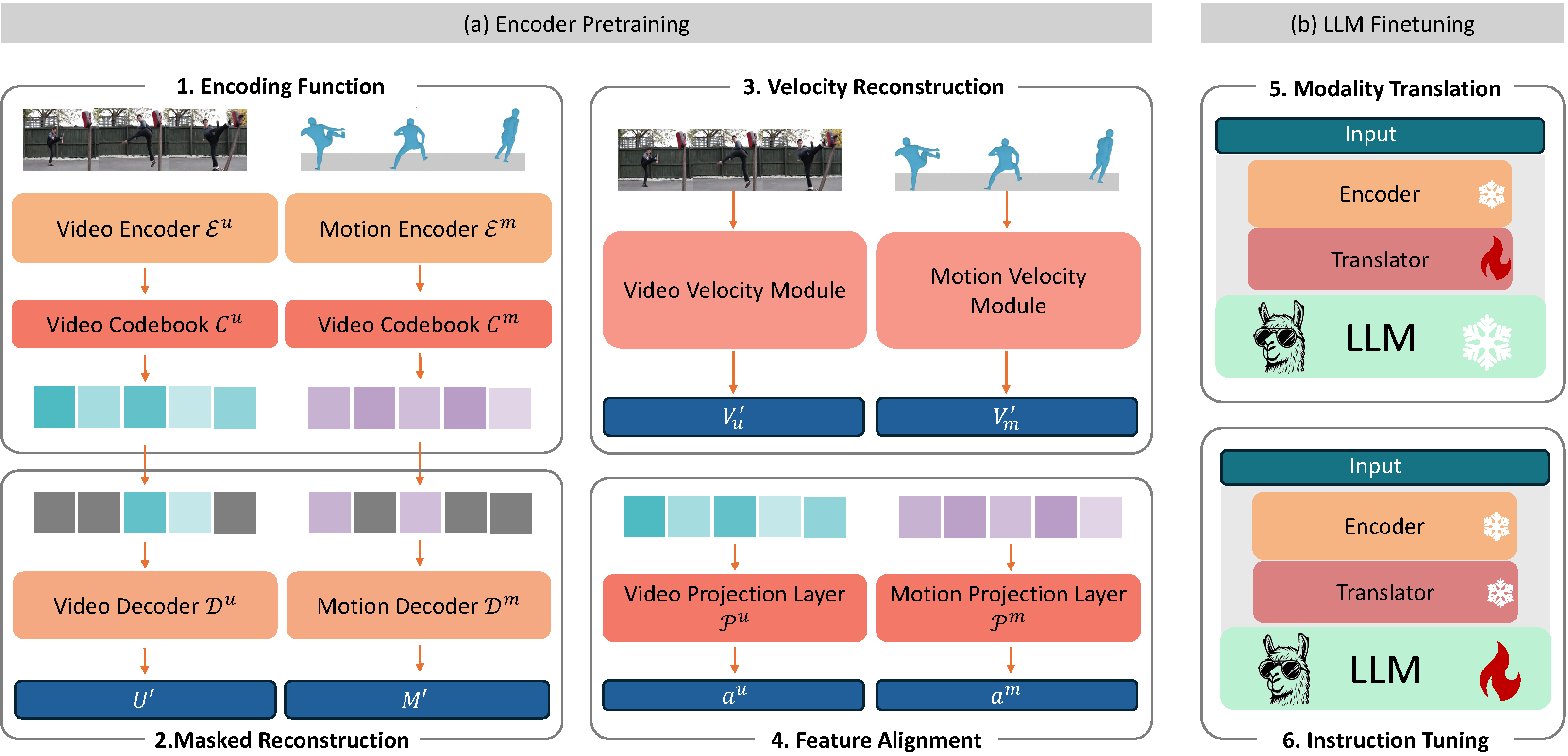}
   \caption{\textbf{System overview of our method.} 
   (a) The encoder pre-training process for learning and aligning video and motion features and enhancing high-frequency details through velocity reconstruction. 
   We utilize a VQ-VAE-based structure, 
   and we design effective learning objectives to enhance the encoder to extract semantic meaningful and fine-grained features.  
   (b) The fine-tuning of the large language model (LLM) for video and motion reasoning consists of two stages: Modality Translation and Multi-modality Instruction Tuning. 
   In the Modality Translation stage, 
   we train a translation layer for each modality to map the encoding feature to the LLM space. 
   In the Instruction Tuning stage, we fine-tune the LLM to understand human motion and videos for downstream tasks.}
   \label{fig:overview}
   \vspace{-10pt}
\end{figure*}


Our objective is to develop 
an LLM-based system capable of understanding and reasoning 
both human video and motion inputs. 
Given a question $Q$ and an input, 
which can be either a video $U$ or motion sequence $M$, 
the model generates the corresponding answer $W$. 
{\it Specifically,} 
a video $U$ is represented as a sequence of $T$ keyframes, 
denoted as $\{u_0, u_1, ..., u_{T-1}\}$. 
Similarly, a motion sequence $M$ is represented as 
$K$ consecutive pose frames, 
denoted as $\{m_0, m_1, ..., m_{K-1}\}$.

We present an overview of the pipeline 
in Fig.~\ref{fig:overview}.
To enable effective multi-modal human motion understanding, 
our framework focuses on: 
(1) discovering human motion concepts from both video and motion data,
(2) preserving high-frequency motion details using velocity reconstruction to reduce temporal smoothing,
and 
(3) aligning video and motion features to enhance multi-modal comprehension. 
To accomplish this, 
we train our video and motion encoders with four key learning objectives: 
\emph{Masked Reconstruction}, \emph{Discriminative Informativeness}, \emph{Actionable Informativeness}, and \emph{Feature Alignment}. 
Consequently, we leverage the encoded features of videos and motions to facilitate the fine-tuning of LLMs for video-motion understanding and reasoning.
The following sections detail each component of our approach, 
including multi-modality encoder pre-training, 
modality translation, 
and instruction tuning.

\subsection{Multi-Modality Encoder Pre-training via\\
Human Motion Concept Discovery}
Pre-training encoders are critical for multi-modality large language models (LLMs)~\cite{wang2022internvideo,wang2024internvideo2,zhu2023languagebind}, 
as they enable the effective extraction of essential features 
from large-scale input data. 
To enhance the encoder's ability to capture key information, 
we design three distinct learning objectives: 
masked reconstruction, velocity reconstruction, and feature alignment, 
all contributing to the quality of the discovered 
human motion concepts as well as the pre-training efficiency.
For both videos and motions, 
we adopt a similar network structure during encoder pre-training. 
Consequently, in the following section, 
we focus on detailing the network structure for video inputs.

\paragraph{Encoding Function.}
In our approach, 
we design masked autoencoder architectures tailored for both video and motion data. 
{\it Specifically,} 
for videos, we utilize a pre-trained encoder $\mathcal{E}_u$, 
to encode the input $U$ into a feature representation $f^u$:
\begin{equation}
    F^u = \mathcal{E}_u(U).
\end{equation}

Then, we employ a codebook $C^u$, 
modeled with VQ-VAE to quantize $f^u$ into a discrete feature representation $d^u$:
\begin{equation}
    D^u = \text{Quantize}(F^u, C^u),
\end{equation}
where `$\text{Quantize}$' denotes the quantization operation, and $C^u$ is the codebook used for discretization, also serving as the discovered human motion concepts.

\paragraph{Masked Reconstruction.}
The masked autoencoder
has proven to be highly effective for pre-training \cite{he2022masked}, 
as it facilitates learning of meaningful and low-frequency features
from input data.
Thus, 
we leverage masked auto-encoding 
to enhance the discovered concepts 
and facilitate the multi-modal feature pre-training. 

More explicitly, 
we mask $D^u$ by a probability of $\sigma$ 
to obtain the masked feature $D^u_{\text{mask}}$:
\begin{equation}
    D^u_{\text{mask}} = \text{Mask}(D^u, \sigma),
\end{equation}
where `$\text{Mask}$' 
is the operation that applies masking to the features, 
and $\sigma<1.0$ controls the masking ratio. 
A video decoder $D^u$ is used to reconstruct 
the video $U'$ from $\mathcal{D}^u_{\text{mask}}$:
\begin{equation}
    U' = \mathcal{D}^u(D^u_{\text{mask}}).
\end{equation}

We compute the reconstruction loss between the original video $U$ and the reconstructed video $U'$ as follows:
\begin{equation}
    \mathcal{L}^{\text{rec}}_{\text{video}} = || U' - U ||^2_2.
\end{equation}

Similarly, 
we introduce a reconstruction loss for motion data, 
$\mathcal{L}^{\text{rec}}_{\text{motion}}$, 
following the same process. 
The total reconstruction loss is defined as:
\begin{equation}
    \mathcal{L}^{\text{rec}}=
    \mathcal{L}^{\text{rec}}_{\text{video}} + \mathcal{L}^{\text{rec}}_{\text{motion}}.
\end{equation}

The above masked reconstruction loss 
ensures that the encoded features contain sufficient information about the video or motion input, 
while learning away possible nuisances through masking and 
making the learning of the concepts (codebook) more focused on critical information.
On the other hand,
the discretization imposed by the concepts 
further enhances the meaningfulness and robustness 
of the pre-trained multi-modal (motion) features.

\paragraph{Velocity Reconstruction.}

Reconstructing inputs 
from masked features inevitably results in the loss of high-frequency details, leading to temporally over-smoothed reconstruction outcomes and preventing the discovery of more nuanced yet useful motion concepts.

To address this limitation, 
we propose an objective that further 
reconstructs 
the ``velocity'' of the input sequence. 
To be more specific, 
we define ``state'' 
as the encoding or feature of a frame in the video or motion sequence, 
while the ``velocity'' means the difference between consecutive ``states'' that captures 
high-frequency information by focusing on dynamic changes. 
For video and motion data, 
we denote the ``states'' as $U$ and $m$, 
and the ``velocity'' as the optical flow $O$ (computed from image frames)
and delta motion $\delta M$, respectively.

\begin{figure*}[htb]
  \centering
  \includegraphics[width=0.8\linewidth]{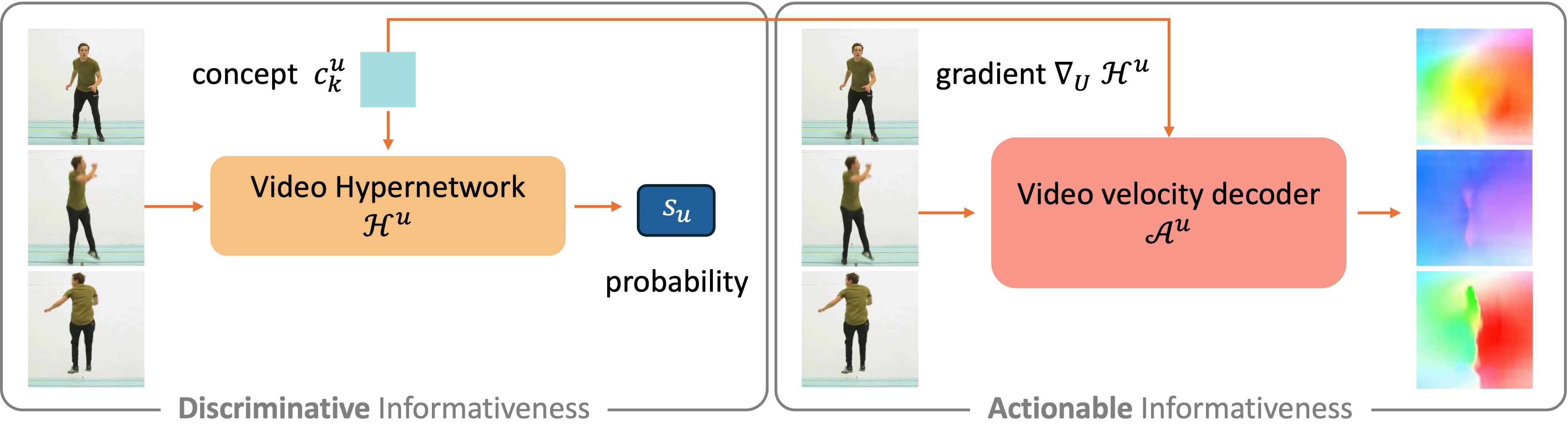}
   \caption{\textbf{Overview of the velocity reconstruction components.} We build similar network structures for both video and motion, and we present the video part in this figure. This module is composed of two learning objectives. 1) Discriminative informativeness (left) aims to improve the distinctiveness of encoded features by reducing representational ambiguity. 2) Actionable informativeness (right) focuses on reconstructing the velocity by leveraging gradient information from the discrimination hypernetwork. As for the video data, we employ optical flow as the representation of the velocity.}
   \label{fig:velocity}
   \vspace{-10pt}
\end{figure*}

{\it However,} 
directly reconstructing velocity 
from the same set of masked features 
introduces difficulty in figuring out the dynamic information, 
given the above-mentioned over-smoothing effect in the encoded features. 
To mitigate this, 
we draw inspiration from the robotic manipulation literature~\cite{liu2024infocon} on action concept discovery 
and incorporate two tailored learning objectives 
-- \emph{discriminative informativeness} 
and \emph{actionable informativeness} -- 
to enhance the efficiency of representations 
and improve the quality of temporal dynamics 
in multi-modality pretaining, 
as shown in Fig.~\ref{fig:velocity}.

{\bf Discriminative informativeness.} 
The discriminative informativeness objective 
aims to improve the distinctiveness of encoded features 
by increasing the correlation between the discovered concepts and the corresponding state features. 
{\it Specifically,} 
we introduce a video hyper-network $\mathcal{H}^u$, 
which takes discrete features $c^u_k$ from the codebook $C^u$ 
as input and generates a classification network. 
The hyper-network $\mathcal{H}^u$ 
(together with the generated classifier)
is trained to determine 
whether an input state 
is compatible with its corresponding code class, 
which is represented by a confidence value $s_u$ (classification score) from 0 to 1, 
thereby enhancing the distinguishability 
and separation of each feature class 
in the representation space.


The discriminative loss for video is defined as:

\begin{equation}
    \mathcal{L}^{\text{dis}}_{\text{video}} = 
    \sum^{|C^u|-1}_{k=0} \sum^{T-1}_{i=0} 
    CE(\mathcal{H}^u(c^u_k, u_i), d^u_i),
\end{equation}
where $|C^u|$ is the size of the codebook $C^u$, 
$CE(\cdot)$ is the cross-entropy loss for classification, 
$d^u_i$ represents the ``ground-truth label'' for the input state $u_i$ determined by the VQ-VAE,
and $T$ is the number of frames. 

{\bf Actionable informativeness.} 
The actionable informativeness objective 
focuses on reconstructing the velocity by leveraging 
gradient information from the discriminative hyper-network, $\mathcal{H}^u$ introduced above. 
Inspired by the idea that 
gradients of a discriminator function 
can inform the state changes, as observed in~\cite{liu2024infocon}, 
this objective ensures that the model captures 
not only static state information, 
but also the dynamic changes in the sequence. 
By leveraging gradient fluctuations around a given state, 
the approach effectively captures 
dynamic motion details, 
such as velocity, resulting in more fine-grained motion concepts, 
and enriching the motion representation.


{\it Specifically,} 
we initialize a network $\mathcal{A}^u$ for video data, 
which takes key image frames $U$ and gradients from the discriminator $\nabla_U \mathcal{H}^u(D^u, U)$ as inputs and predicts the velocity $V_u'$. 
The  actionable informativeness loss for video is then computed as: 
\begin{equation}
\mathcal{L}^{\text{act}}_{\text{video}} = ||V_u - V_u'||^2_2.
\end{equation}
where $V_u$ represents the ground-truth velocity of video frames. 
This process is repeated for motion data to compute the corresponding loss $\mathcal{L}^{\text{act}}_{\text{motion}}$ using $V_m$ and $V_m'$.

{\bf Combined Objectives.} 
By integrating both video and motion components, 
we derive the total velocity reconstruction loss for human motion concept discovery and motion feature pre-training 
as in the following:
\begin{equation}
    \mathcal{L}^{\text{dis}} = \mathcal{L}^{\text{dis}}_{\text{video}} + \mathcal{L}^{\text{dis}}_{\text{motion}}
\end{equation}
\begin{equation}
    \mathcal{L}^{\text{act}} = \mathcal{L}^{\text{act}}_{\text{video}} + \mathcal{L}^{\text{act}}_{\text{motion}}   
\end{equation}

With the above objectives, 
our approach ensures that the model 
effectively reconstructs high-frequency details, 
improving the temporal fidelity and 
overall quality of multi-modal representations. 


\paragraph{Feature Alignment.}
Multi-modality models 
often leverage the shared information 
across different modalities to enhance performance \cite{zhu2023languagebind,chen2024motionllm,rajasegaran2023benefits}. 
In the case of motion and video, 
video data provides contextual information that 
motion data alone may be lacking, 
which is critical for accurate motion prediction and understanding. 
Conversely, 
motion data focuses on human-centric features in the video, 
reducing over-reliance on environmental details. 
An efficient utilization or balancing 
of both is beneficial 
for the human motion LLMs.
{\it However,} 
prior work on human motion and video understanding \cite{chen2024motionllm} 
only achieves implicit alignment through paired data 
during LLM fine-tuning, 
without explicit alignment in encoder training. 
To address this issue, 
we propose an alignment loss to improve information sharing and cross-modal coherence.

To achieve alignment between video and motion features, 
we collect a set of video-motion pairs from Motion-X \cite{lin2024motion}, 
which are aligned by frames.
We instantiate two projection layers: 
The video projection layer $\mathcal{P}^u$ 
and the motion projection layer $\mathcal{P}^m$. 
These layers are designed to map the respective features 
from the video and motion domains into a shared aligned space.

More explicitly, 
given paired video and motion data $\{U, M\}$, 
we use $\mathcal{P}^u$ and $\mathcal{P}^m$ to project the discrete video feature $d^u_i$ and motion feature $d^m_j$ into the aligned feature space, resulting in $a^u_i$ and $a^m_j$, where 
$a^u_i, a^m_j \in \mathbb{R}^H$. 
Here, $H$ denotes the dimensionality of the aligned features in the shared space.

To ensure effective alignment, 
we define an alignment loss as follows:
\begin{equation}
    \mathcal{L}^{\text{align}}= 
    \sum^{T-1}_0 
    \frac{\text{norm}(a^u_i) \cdot \text{norm}(a^m_{\text{align}(i)}) / \epsilon}{\sum^{K-1}_0 \text{norm}(a^u_i) \cdot \text{norm}(a^m_j) / \epsilon},
\end{equation}
where $\text{align}(i)$ indicates the pose index 
in the motion sequence corresponding to the key video frame $u_i$, 
$\text{norm}$ represents a normalization operation to stabilize feature magnitudes and $\epsilon$ is a temperature parameter that controls the sharpness of the alignment distribution. 

Note that $\mathcal{L}^{\text{align}}$ 
is computed only when motion and video are paired. 
For standalone motion or video inputs, 
this loss term is omitted for applicability.


The {\bf overall} 
human motion concept discovery objective 
is defined as:
\begin{align}
    \mathcal{L}^{\text{total}} = &
    \mathcal{L}^{\text{rec}}_{\text{motion}} + \mathcal{L}^{\text{rec}}_{\text{video}} + \nonumber \\
    & \lambda^{\text{dis}} \mathcal{L}^{\text{dis}} + \lambda^{\text{act}} \mathcal{L}^{\text{act}} + \lambda^{\text{align}} \mathcal{L}^{\text{align}},
\end{align}
where $\lambda^{\text{dis}}$, $\lambda^{\text{act}}$ and $\lambda^{\text{align}}$ are weighting coefficients for the respective losses $\mathcal{L}^{\text{dis}}$, $\mathcal{L}^{\text{act}}$ and $\mathcal{L}^{\text{align}}$.

\subsection{Modality Translation}
Although pre-trained encoders 
capture rich features from the original input, 
these features are often difficult for LLMs to interpret directly \cite{chen2024motionllm,lin2023video,li2024llava}. 
{\it Therefore,} 
it is essential to introduce an additional projection layer 
to map these features into a space that LLMs can effectively understand.

As illustrated in Fig~\ref{fig:overview}, 
we add tunable translation layers after the encoders $\mathcal{E}^u$ and $\mathcal{E}^m$. 
These layers are trained using annotated video and motion sequence data to learn task-specific transformations. 
During this stage, the encoders and the LLM $\mathcal{F}$ are kept frozen to prevent adverse feature drift, 
and only the weights of the translation layers are updated. 
This strategy preserves the pre-trained knowledge of the encoders and language model while optimizing the feature mapping for downstream tasks.

\subsection{Multi-Modality Instruction Tuning}
Finally, 
we employ instruction tuning 
to train the language model $\mathcal{F}$ to respond to user queries using the corresponding video or motion sequences 
as context. 
During this process, 
the encoders and projection layers remain frozen, 
ensuring their pre-trained features are retained, 
while the weights of the language model $\mathcal{F}$ are updated to accommodate the task.

In this stage, 
the model can further enhance cross-modal alignment 
using paired motion-video data in an implicit manner. 
To streamline the fine-tuning process, 
we adopt LoRA \cite{hu2021lora} with a rank of 8, 
allowing efficient and lightweight adjustment to the language model.

\begin{table*}[!t]
\centering
\caption{ {\bf Comparison of different methods on BABEL-QA test set.} The pred type represent the prediction type, including classification and generaration. For 2s-AGCN and MotionCLIP, 'M' means MLP, 'R' means RNN. Our method outperforms the baselines across all the metrics.}
\label{tab:babel}
\begin{tabular}{ccccccccc}
\hline
\multirow{2}{*}{Model} & \multirow{2}{*}{Pred type} & \multirow{2}{*}{Overall} & \multicolumn{3}{c}{Query type} & \multicolumn{3}{c}{Relation Type} \\ \cline{4-9} 
 &  &  & Action & Direction & BodyPart & Before & After & Other \\ \hline
2s-AGCN-M & cls. & 0.355 & 0.384 & 0.352 & 0.228 & 0.331 & 0.264 & 0.295 \\
2s-AGCN-R & cls. & 0.357 & 0.396 & 0.352 & 0.194 & 0.337 & 0.301 & 0.285 \\
MotionCLIP-M & cls. & 0.430 & 0.485 & 0.361 & 0.272 & 0.372 & 0.321 & 0.404 \\
MotionCLIP-R & cls. & 0.420 & 0.489 & 0.310 & 0.250 & 0.398 & 0.314 & 0.387 \\ 
MotionLLM & gen. & 0.436 & 0.517 & 0.354 & 0.154 & 0.427 & 0.368 & 0.529 \\
\hline
Ours & gen. & \textbf{0.711} & \textbf{0.809} & \textbf{0.697} & \textbf{0.623} & \textbf{0.707} & \textbf{0.635} & \textbf{0.797} \\ \hline
\end{tabular}
\vspace{-10pt}
\end{table*}

\section{Experiments}
\subsection{Experiments Settings}
\paragraph{Implementation details}
For videos and motions, we initialize a codebook with 512 codes, a one-layer projector for feature alignment, and a 2-layer MLP to generate the weights for a 2-layer discriminator separately. For video inputs, we utilize pre-trained SigLIP~\cite{zhai2023sigmoid} with an image patch size of 16 pixels and hidden dimension of 256 as the video encoder, and we take a 12-layer transformer as the reconstruction decoder. A 4-layer transformer is utilized as the motion encoder and a 2-layer transformer is used for motion reconstruction. 
For the velocity reconstruction, we initialize a two-layer hypernet and two-layer velocity decoder. Besides, we utilize one-layer MLPs for feature alignment.
We pre-train the motion encoder with only the reconstruction loss for 60K iterations with a learning rate of 1e-4. Then, we train both the video and motion encoders with the total loss for 8K iterations with a learning rate of 1e-5.

We utilize Llama3.1-8B as our base LLM model. For the modality translation stage, we initialize a one-layer projector for the video and motion encoder separately. We train the projectors on the annotation dataset for one batch with a batch size of 32 and a learning rate of 1e-4. In the Multi-modality instruction tuning stage, we train the LLM model with a batch size of 16 and a learning rate of 3e-4.

\paragraph{Training datasets}
For the encoder pre-training stage, we utilize Motion-X~\cite{lin2024motion} datasets which contain 81K motion sequences and 36K motion-video pairs. 
In order to facilitate the training process of LLMs, we follow MotionLLM~\cite{chen2024motionllm} to prompt LLMs to generate 20K annotation for motion-video pairs from the Motion-X dataset, and 10K annotation data from the HumanML3D~\cite{Guo_2022_CVPR} dataset. This data is utilized for modality translation. We additionally utilize the VATEX~\cite{wang2019vatex} dataset which contains 20K annotated videos for video translator training. 
For multi-modality instruction tuning, we follow MotionLLM to generate 200K Q\&A pairs for both motions and videos. We also utilize the Video-ChatGPT~\cite{maaz2023video} and BABEL-QA~\cite{endo2023motion} datasets to enhance the motion and video understanding ability of our model, which has 100K and 2K Q\&A data separately.  

\paragraph{Evaluation datasets}
Although MotionLLM~\cite{chen2024motionllm} creates a large benchmark for both motion and video understanding, by the time we completed our paper, they had not released the specific details or access methods for the benchmark, nor did they provide a clear timeline for its release. Therefore, we did not perform comparisons on that benchmark but instead selected the test set of ActivityNet-QA~\cite{yu2019activityqa} and BABEL-QA~\cite{endo2023motion} to separately validate the performance of our algorithm on video and motion data.

\paragraph{Evaluation metrics}
On the BABEL-QA benchmark, we evaluate the prediction accuracy for evaluation. As for the classification-based baselines, we evaluate the accuracy directly. As for the generative baselines, we adapt the evaluation protocol used in previous works~\cite{chen2024motionllm,li2024llava} by utilizing LLMs. Specifically, the evaluation LLM is given some examples of questions, answers, and scores (from 0 to 1), and it is required to follow these examples and score the experiment results. 
On the Activity-QA benchmark, we only compare our methods with our generative baselines. Thus, we directly utilize LLMs to evaluate the methods.

\subsection{Quantitative Results}

\paragraph{Experiments on ActivityNet-QA.}

\begin{table}[b]
\centering
\caption{\textbf{Experiments results on ActivityNet-QA benchmark.} }
\label{tab:activity}
\begin{tabular}{cccc}
\hline
Model & \multicolumn{1}{l}{{}} & Acc$\uparrow$ & Score $\uparrow$ \\ \hline
FrozenBiLM~\cite{salewski2023zero} &  & 24.7 & - \\
VideoChat~\cite{li2023videochat} &  & - & 2.2 \\
LLaMA-Adapter~\cite{zhang2023llama} &  & 34.2 & 2.7 \\
Video-LLaMA~\cite{zhang2023video} &  & 12.4 & 1.1 \\
Video-ChatGPT~\cite{maaz2023video} &  & 35.2 & 2.7 \\
Video-LLaVA~\cite{lin2023video} &  & 45.3 & 3.3 \\
VideoChat2~\cite{li2024mvbench} &  & 49.1 & 3.3 \\
MotionLLM~\cite{chen2024motionllm} &  & 53.3 & 3.5 \\ \hline
Ours &  & \textbf{54.2} & \textbf{3.6} \\ \hline
\end{tabular}
\end{table}

\begin{table*}[htb]
\centering
\caption{ Ablation study on BABEL-QA test set. }
\label{tab:ablation}
\begin{tabular}{ccccccccc}
\hline
\multirow{2}{*}{Model} & \multirow{2}{*}{Pred type} & \multirow{2}{*}{Overall} & \multicolumn{3}{c}{Query type} & \multicolumn{3}{c}{Relation Type} \\ \cline{4-9} 
 &  &  & Action & Direction & BodyPart & Before & After & Other \\ \hline
MotionLLM & gen. & 0.436 & 0.517 & 0.354 & 0.154 & 0.427 & 0.368 & 0.529 \\ \hline
Ours-w/o$\mathcal{L}^{rec}$ & gen. & 0.696  & 0.741 &  0.645  & 0.577 & 0.600 & 0.597 & 0.762 \\
Ours-w/o$\mathcal{L}^{dis} \& \mathcal{L}^{act}$ & gen. & 0.637 & 0.693 & 0.478 & 0.606 & 0.667 & 0.526 & 0.709 \\
Ours-w/o$\mathcal{L}^{align}$ & gen. & 0.675 & 0.743 & 0.579 & 0.523 & 0.584 & 0.570 & 0.743 \\
\hline
Ours & gen. & \textbf{0.711} & \textbf{0.809} & \textbf{0.697} & \textbf{0.623} & \textbf{0.707} & \textbf{0.635} & \textbf{0.797} \\ \hline
\end{tabular}
\vspace{-8pt}
\end{table*}

We evaluate the video understanding ability of our method and other baselines on the ActivityNet-QA benchmark as shown in Tab.~\ref{tab:activity}. We follow previous methods~\cite{chen2024motionllm,jin2024chat,li2023videochat} to report the accuracy and scores that are evaluated by LLMs. We show in Tab.~\ref{tab:activity} that our method outperforms all the baselines on both metrics. Compared to the video-only method, our method aligns the video feature with the motion feature, making the encoder concentrate on human-centric information. Compared to MotionLLM, our encoder pre-training helps the network to learn more fine-grained information through velocity reconstruction.

\paragraph{Experiments on BABEL-QA.}

We evaluate the motion understanding ability of our method as compared to existing studies on the BABEL-QA~\cite{endo2023motion} dataset. Following~\cite{chen2024motionllm,endo2023motion}, we compare our method with a diverse set of baselines: 1) {\bf 2s-AGCN}~\cite{shi2019two}, a method using 2s-GCN to extract motion features and classify motion by MLP or RNN;
2) {\bf MotionCLIP}~\cite{tevet2022motionclip}, a method that aligns the features of motion and text using CLIP~\cite{radford2021learning} and predicts motion class by MLP or RNN; 3) {\bf MotionLLM}~\cite{chen2024motionllm}, an LLM-based method which has motion reasoning ability.
We follow~\cite{chen2024motionllm} to fine-tune our method with the training set of BABEL-QA and evaluate our method with LLMs.

From Tab.~\ref{tab:babel}, our method outperforms other methods in all metrics by a large margin of nearly 60\% overall. In the ``bodypart" query task, our algorithm achieves 2.29 times the score of the second-place method, MotionCLIP-M (0.623 vs. 0.272). This advantage stems partly from our velocity-based encoder training, which, compared to traditional motion feature extraction, helps us better capture high-frequency information and ensures the learning of fine-grained features. Additionally, unlike MotionLLM, we perform alignment at the encoder pre-training stage, ensuring that the features learned by the encoder are inherently more semantically rich.

\paragraph{Ablation study.}
We present the ablation results in Tab.~\ref{tab:ablation}. We conducted ablation studies on masked reconstruction, velocity reconstruction, and feature alignment. MotionLLM serves as an ablation baseline without any of these techniques. As presented, the absence of any single module leads to a decrease in performance. Among these, masked reconstruction has the least impact, as we initially train the motion encoder solely with masked reconstruction loss to ensure stable subsequent training. Removing velocity reconstruction has the greatest impact on pexrformance since velocity provides high-frequency information that other modules lack. Likewise, removing the alignment loss significantly affects results, highlighting the importance of multi-modality alignment in motion understanding.

\begin{figure*}[htbp]
  \centering
  \includegraphics[width=0.88\linewidth]{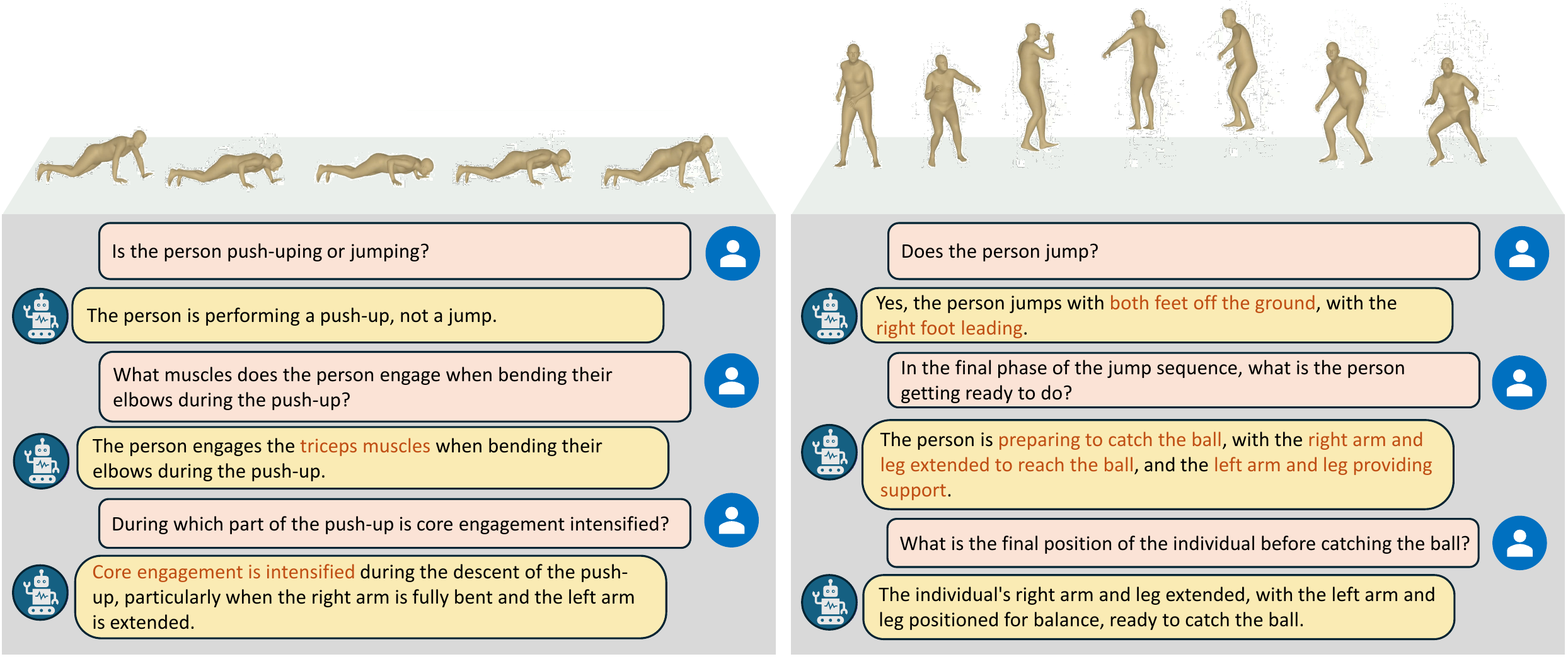}
   \caption{\textbf{Example of motion understanding.} We validate the proposed model’s ability to understand motion from multiple aspects. Q1 on the left and Q1--3 on the right evaluate its comprehension of motion sequences, while Q2 and Q3 on the left push it to analyze kinematic, kinesthetic, and physical properties.
   These results demonstrate our algorithm's ability to recognize, describe in detail, and analyze motions.}
   \label{fig:motionvis}
   \vspace{-11pt}
\end{figure*}

\subsection{Qualitative Results}
To validate our algorithm's ability to understand motions and its contextual knowledge, we conducted multiple rounds of Q\&A with the motion model.
We present two examples in Fig.~\ref{fig:motionvis}. We discover that, beyond its basic motion recognition capabilities, our algorithm can provide detailed descriptions of motions. In the example on the right, it can precisely describe the actions of the left hand and foot, as well as the right hand and foot. Our algorithm also possesses extensive knowledge about various movements. For instance, when asked, ``what muscles does the person engage when bending their elbows during a push-up?" it can accurately respond with ``triceps muscles."
This demonstrates that our algorithm can understand the input motion sequences and perform corresponding analyses.
We present more qualitative results in the supplementary material.

\section{Conclusion}
In this paper, we presented HuMoCon, a novel motion-video understanding framework for human action understanding and concept discovery.
We demonstrated the effectiveness of our approach through comprehensive experiments, showcasing major improvements over state-of-the-art methods. Our work provides a strong foundation for future research in human action understanding and motion concept discovery. 
There are still some drawbacks of our work, such as the lack of human-environment contact or understanding niche motions.
We plan to address those limitations by expanding the framework to incorporate additional modalities, particularly as environments, objects, and contact, and curate a wider range of specialized datasets.

\clearpage
\subsection*{Acknowledgment}
This work is supported by 
the Early Career Scheme of the Research Grants Council 
(grant \# 27207224),
the HKU-100 Award, 
a donation from the Musketeers Foundation, and an Academic Gift
from Meta. The data collection and processing and model developing was done at the the the University of Hong Kong.

{
    \small
    \bibliographystyle{ieeenat_fullname}
    \bibliography{main}
}

\clearpage
\setcounter{page}{1}
\maketitlesupplementary

\begin{table*}[hbp]
\centering
\caption{The dataset details for motion data. Motion-XQA* refers to the training dataset generated following the method of Motion-XQA~\cite{chen2024motionllm}.}
\label{tab:datam}
\begin{tabular}{clclcclcc}
\hline
         &  & Encoder Pretraining &  & \multicolumn{2}{c}{Modality Translation} &  & \multicolumn{2}{c}{Instruction Tuning} \\ \hline
Data     &  & Motion-X&  & Motion-X caption       & Humanml3D       &  & Motion-XQA          & BABEL-QA         \\
\# count &  & 81 K    &  & 20 K       & 10 K&  & 200 K   & 2 K  \\ \hline
\end{tabular}
\end{table*}

\begin{table*}[hbp]
\centering
\caption{The dataset details for video data. Motion-XQA* refers to the training dataset genarated following the method of Motion-XQA~\cite{chen2024motionllm}.}
\label{tab:datav}
\begin{tabular}{clclcclcc}
\hline
         &  & \multicolumn{2}{c}{Modality Translation} &  & \multicolumn{2}{c}{Instruction Tuning} \\ \hline
Data     &  & Motion-X caption         & VATEX         &  & Motion-XQA*       & Video-ChatGPT      \\
\# count &  & 20 K         & 20 K          &  & 200 K & 100 K  \\ \hline
\end{tabular}
\end{table*}

\section{Implementation Details}
We resize all video frames to a resolution of 320 × 256. Limited by GPU memory, we only take 8 key frames for each video following MotionLLM~\cite{chen2024motionllm}. And we represent the motion following Humanml3D~\cite{guo2022generating}.

During the encoder pre-training stage, we set the mask ratio $\lambda$ to 0.75 and configure the loss weights $\lambda^{dis}$, $\lambda^{act}$ and $\lambda^{align}$ to 0.3, 0.1, and 0.1, respectively. To stabilize the training of VQ-VAE, we apply the Exponential Moving Average (EMA) to update the codebook. 
The encoder is trained using one A100 GPU with a micro-batch size of 16, and gradients are accumulated over 8 steps before performing backpropagation.

For modality translation, we train the motion and video projection layers using a single A100  GPU with a batch size of 32. The LLMs are further trained with a batch size of 16. To align the output template of BABEL-QA~\cite{endo2023motion}, we fine-tune the LLM model on BABEL-QA training set using a  batch size of 16. We also list the dataset details in Tab.~\ref{tab:datam} and Tab.~\ref{tab:datav}.




\section{Additional Visualizations}

\subsection{Visualizations for Motion Understanding}
\begin{figure*}[]
  \centering
  \includegraphics[width=1\linewidth]{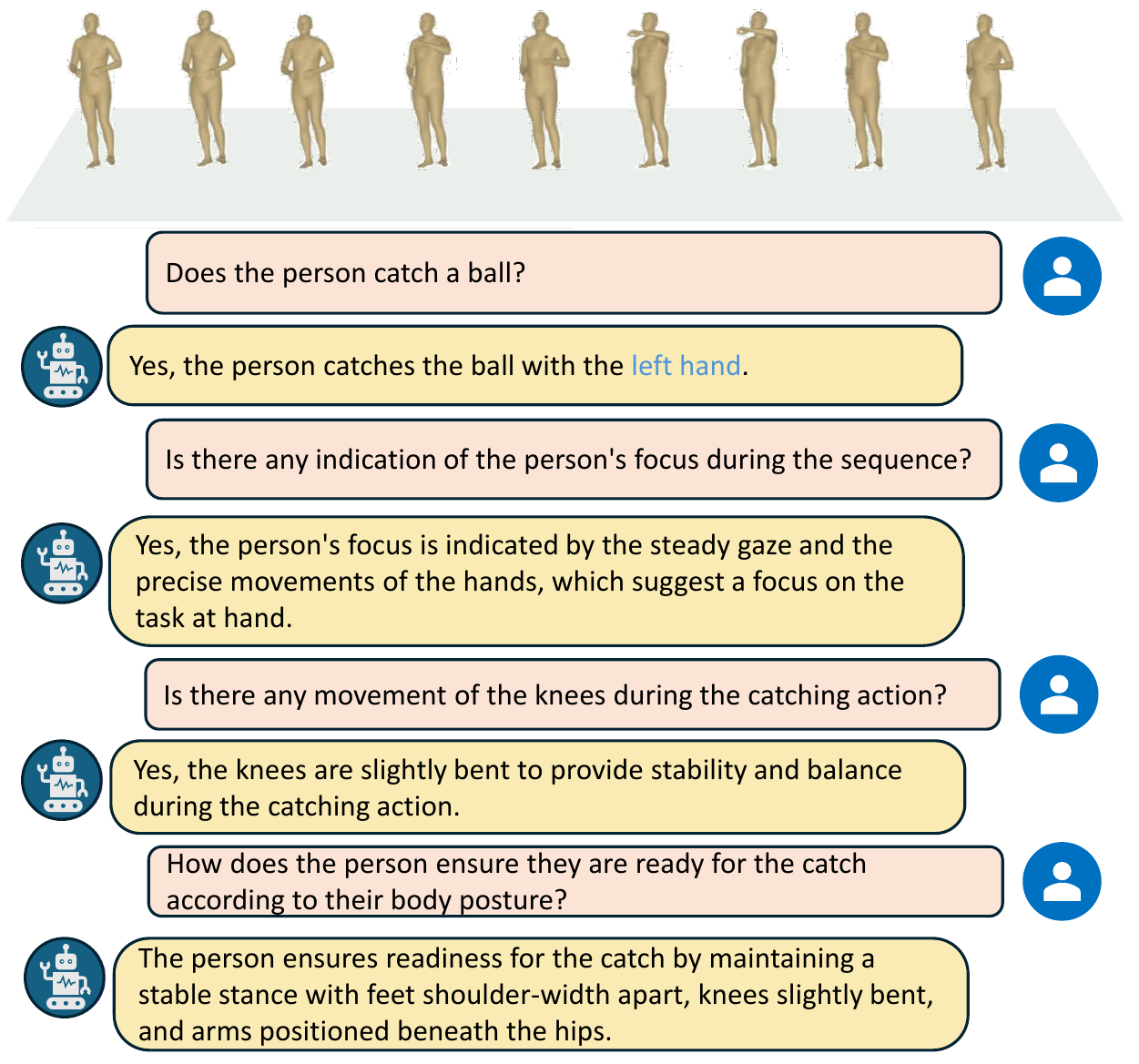}
   \caption{\textbf{Visualizations for motion understanding.}}
   \label{fig:motionvis2}
\end{figure*}

\begin{figure*}[]
  \centering
  \includegraphics[width=1\linewidth]{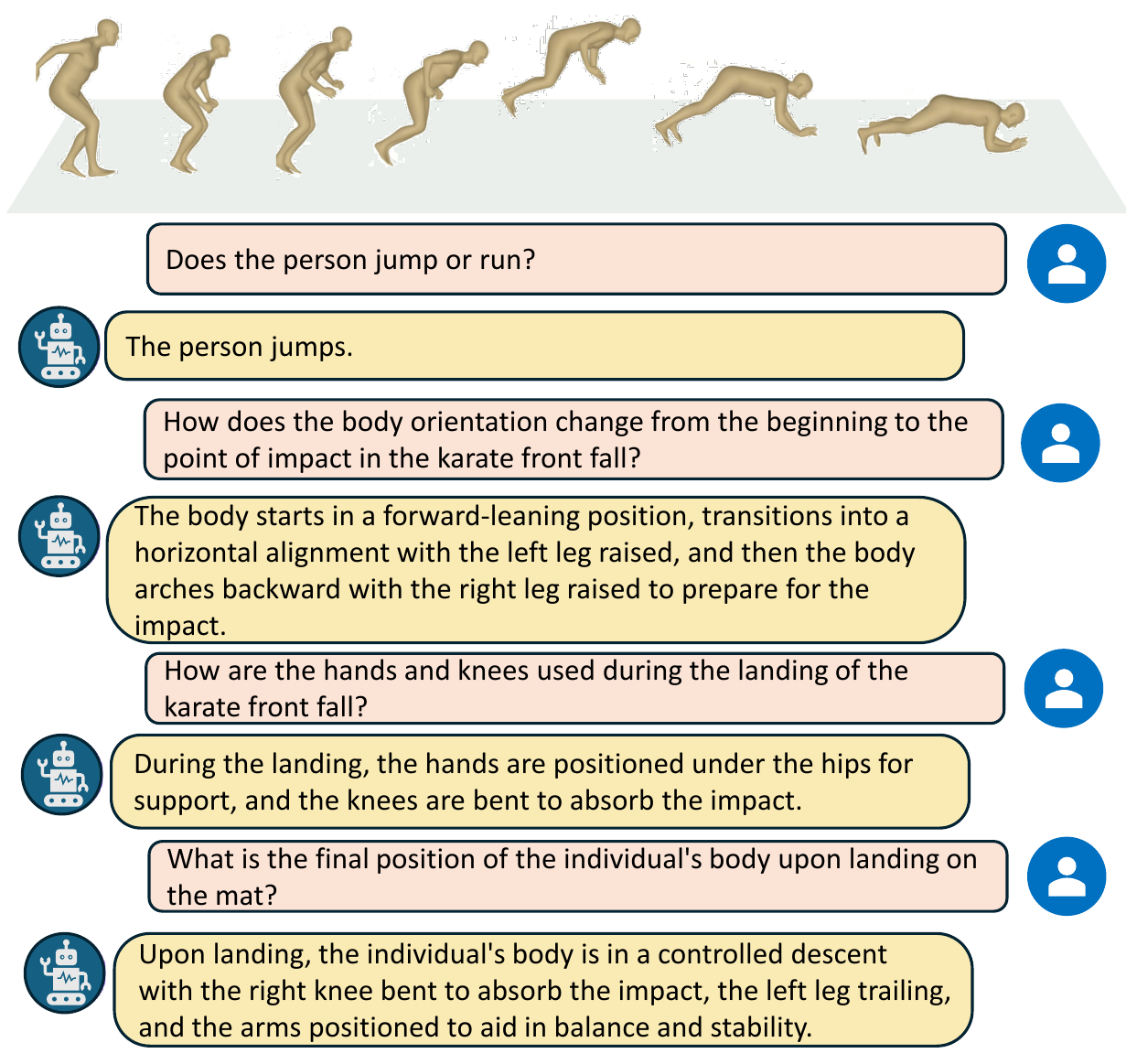}
   \caption{\textbf{Visualizations for motion understanding.}}
   \label{fig:motionvis3}
\end{figure*}

Additional visualization results for motion understanding are provided in Fig.~\ref{fig:motionvis2} and Fig.~\ref{fig:motionvis3}. 
These examples highlight our method's ability to capture detailed motion information and perform accurate motion analyses.

\subsection{Visualizations for Video Understanding}
We also present visualizations for video understanding in Fig.~\ref{fig:vcomp}.

\subsection{Qualitative Comparisons with Other Video Understanding Methods}
\begin{figure*}[htbp]
  \centering
  \includegraphics[width=1\linewidth]{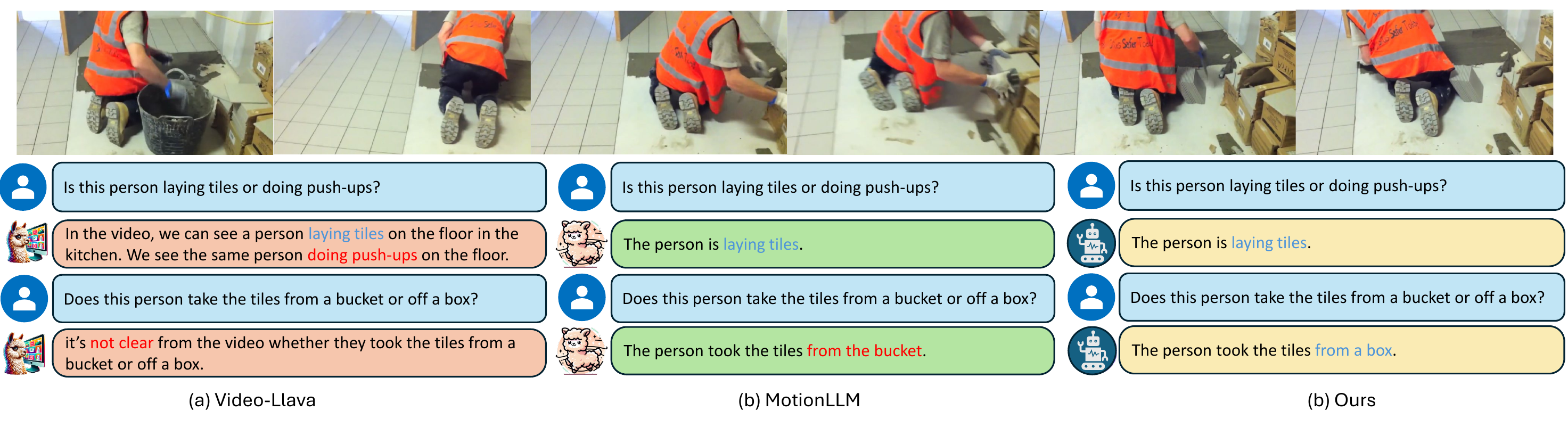}
   \caption{\textbf{Comparison between different video understanding methods.} We compare our method with Video-Llava and MotionLLM. We highlight the correct and wrong answers in blue and red, respectively. }
   \label{fig:vcomp}
\end{figure*}

We compared our algorithm with Video-LLaVA~\cite{zhang2023video} and MotionLLM~\cite{chen2024motionllm} to evaluate their ability to understand video content. As shown in the Fig.~\ref{fig:vcomp}, our algorithm demonstrates a better understanding of the video's content. Video-LLaVA, lacking a deeper comprehension of motion, it struggles to effectively distinguish between actions such as tiling and push-ups. Additionally, Video-LLaVA cannot identify the exact moment when the worker picks up the tile, preventing it from answering questions about where the tiles were picked up.

MotionLLM, on the other hand, leverages motion data to accurately differentiate the workers' actions. However, due to insufficient alignment between motion and video data, it also fails to correctly answer the question regarding where the tile was picked up. In comparison, our algorithm effectively integrates motion and video data, enabling it to address both questions.

\begin{figure}[hp]
  \centering
  \includegraphics[width=1\linewidth]{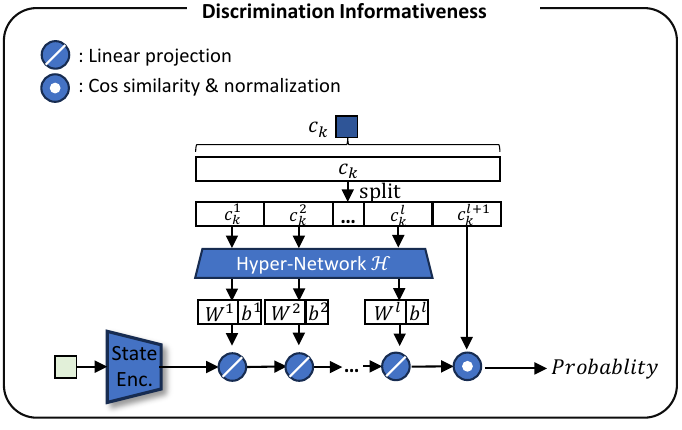}
   \caption{\textbf{The structure of the hyper-network.}}
   \label{fig:hyper}
\end{figure}

\begin{table*}[htbp]
\caption{The network structure of encoder pretraining components}
\centering
\label{tab:net}
\begin{tabular}{cc}
\hline
Components        & Architecture         \\ \hline
\multirow{7}{*}{Motion Encoder $\mathcal{E}^m$} & (0): Linear(263, 256)\\
& (1): ReLU()\\
& (2): Linear(256, 512)\\
& (3): 4 $\times$ TransformerLayer(512, n\_head=(8))   \\
& (4): Linear(512, 256)\\
& (5): ReLU()\\
& (6): Linear(256, 128)\\ \hline
Motion Codebook $C^m$      & (0): Parameter((512, 128), requires\_grad=False) \\ \hline
Motion Projection Layer $\mathcal{P}^m$         & (0): Linear(128, 256)\\ \hline
\multirow{10}{*}{Motion Decoder $\mathcal{D}^m$}       & (0): Linear(263, 256)\\
& (1): ReLU()\\
& (2): Linear(256, 512)\\
& (3): 4 $\times$ CasualTransformerLayer(512, n\_head=(8))   \\
& (4): Linear(512, 256)\\
& (5): ReLU()\\
& (6): Linear(256, 512)\\
& (7): Linear(512, 256)\\
& (8): ReLU() \\
& (9): Linear(256, 263) \\ \hline
\multirow{10}{*}{Motion Velocity Decoder}       & (0): Linear(391, 256)\\
& (1): ReLU()\\
& (2): Linear(256, 512)\\
& (3): 4 $\times$ CasualTransformerLayer(512, n\_head=(8))   \\
& (4): Linear(512, 256)\\
& (5): ReLU()\\
& (6): Linear(256, 512)\\
& (7): Linear(512, 256)\\
& (8): ReLU() \\
& (9): Linear(256, 263) \\

\hline
Video Encoder $\mathcal{E}^u$        & (0): Conv2d(3, 768, kernel\_size=(16, 16), stride=(16, 16)) \\
       & (1): Embedding(256, 768)      \\
       & (2): 12 $\times$ TransformerLayer(768)  \\
       & (3): LayerNorm(768) \\
       & (4): TransformerLayer(768)    \\ \hline
Vision Codebook $C^u$      & (0): Parameter((512, 768), requires\_grad=False)  \\ \hline
Vision Projection Layer $\mathcal{P}^u$        & (0): Linear(768, 256)         \\ \hline
\multirow{3}{*}{Video Decoder $\mathcal{D}^u$} & (0): Linear(768, 384)         \\
       & (1): 10 $\times$ CasualTransformerLayer(384, n\_head=(16))  \\
       & (2): Linear(384, 1536)        \\ \hline
Video Velocity Encoder $\mathcal{E}^u$        & (0): Conv2d(771, 768, kernel\_size=(16, 16), stride=(16, 16)) \\
       & (1): Embedding(256, 768)      \\
       & (2): 12 $\times$ TransformerLayer(768)  \\
       & (3): LayerNorm(768) \\
       & (4): TransformerLayer(768)    \\ \hline
Vision Codebook $C^u$      & (0): Parameter((512, 768), requires\_grad=False)  \\ \hline
Vision Projection Layer $\mathcal{P}^u$        & (0): Linear(768, 256)         \\ \hline
\multirow{3}{*}{Video Decoder $\mathcal{D}^u$} & (0): Linear(768, 384)         \\
       & (1): 10 $\times$ CasualTransformerLayer(384, n\_head=(16))  \\
       & (2): Linear(384, 1536)        \\ \hline

\end{tabular}
\end{table*}

\section{Evaluation Details}
\subsection{Evaluation on BABEL-QA Benchmark}
\begin{figure*}[]
  \centering
  \includegraphics[width=1\linewidth]{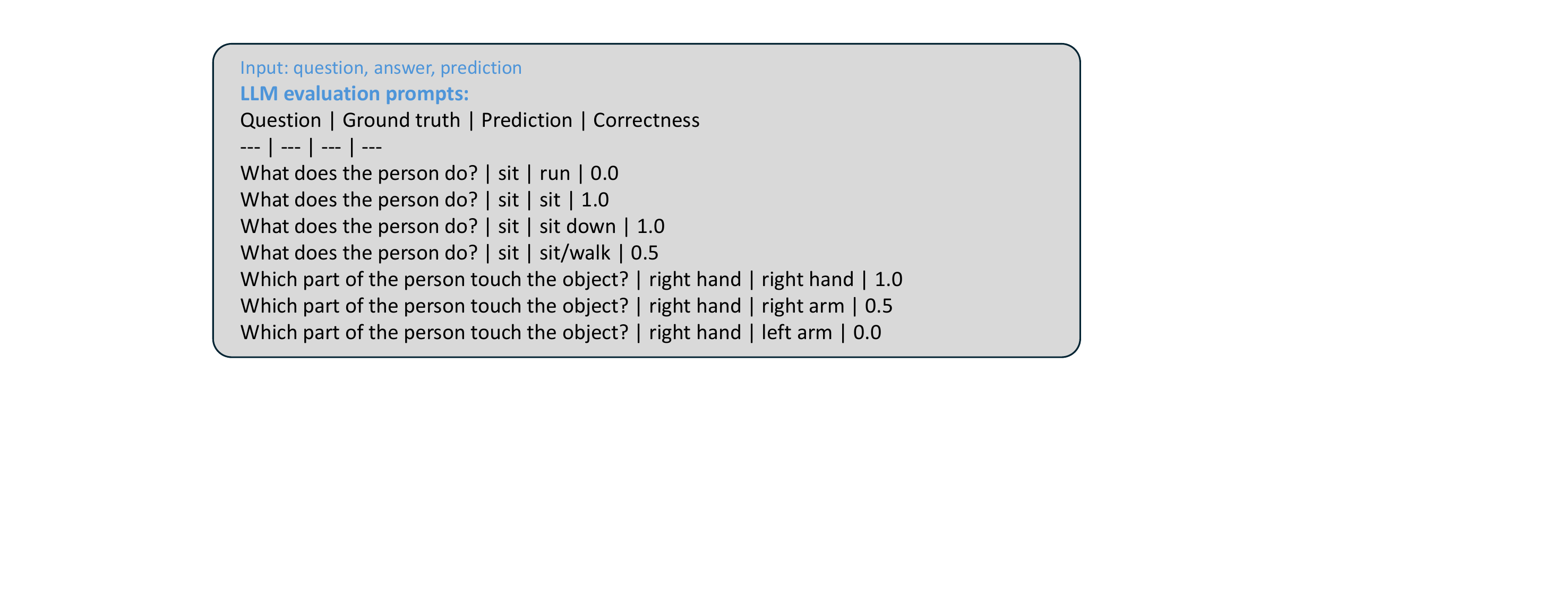}
   \caption{\textbf{Evaluation prompt for BABEL-QA benchmark.}}
   \label{fig:bqa}
\end{figure*}

For the BABEL-QA benchmark, we extend the evaluation protocol used in previous multi-modality LLMs evaluations~\cite{zhang2023video}. The evaluation involves scoring the correctness of each Q\&A pair generated by the LLMs. Detailed information about the evaluation prompt is provided in Fig.~\ref{fig:bqa}.

\subsection{Evaluation on ActivityNet-QA Benchmark}
\begin{figure*}[]
  \centering
  \includegraphics[width=1\linewidth]{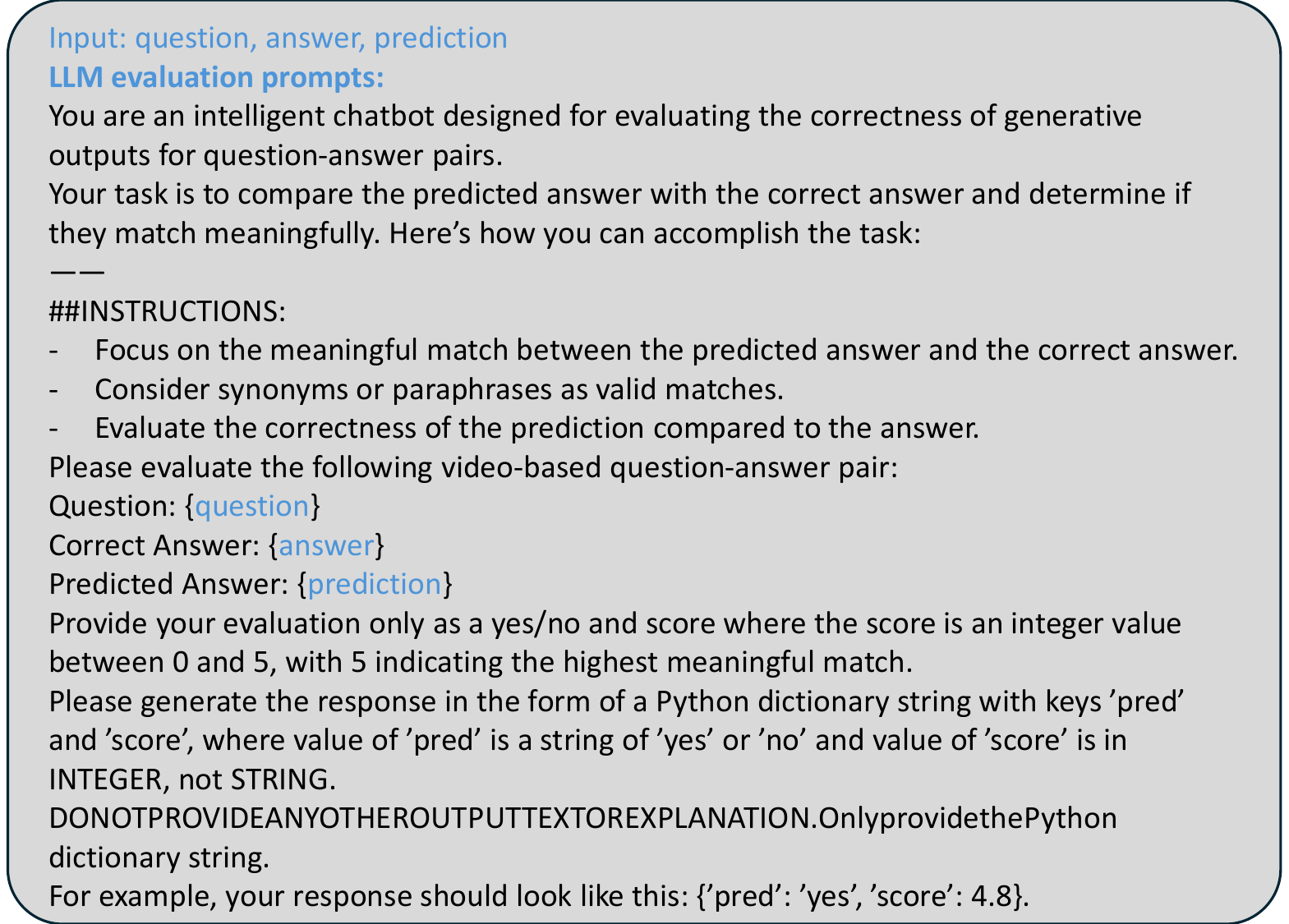}
   \caption{\textbf{Evaluation prompt for ActivityNet-QA benchmark.}}
   \label{fig:ana}
\end{figure*}

On the ActivityNet-QA benchmark, we follow the approach of MotionLLM~\cite{chen2024motionllm} to evaluate both th correctness of the predicted answers and assign a score ranging from 0 to 5, with 5 representing the highest level of semantic match. The details of the evaluation prompt are provided in Fig.~\ref{fig:ana}.

\section{Network Structure for Encoder Pretraining}
\subsection{Structure of Hyper-Networks}
We utilize hyper-networks $\mathcal{H}^u$ and $\mathcal{H}^m$, for velocity reconstruction, following InfoCon~\cite{liu2024infocon}. We split the discrete features $D^u$ and $D^m$ into $K$ parts. We generate $K-1$ layer-network weights using the first $K-1$ feature parts, and the last feature part to calculate the cosine similarity with the output of a 3-layer-network, which is then used as the probability. We present the structure of the hyper-network in Fig.~\ref{fig:hyper}.

\subsection{Structure of Network Components}
We present the structure of the components of our network in Tab.~\ref{tab:net}.

\subsection{Selection of modality}
We set aside a portion of the data as a validation set and evaluated cross-entropy loss using video and motion as inputs. 
The results showed a loss of 0.88 for motion and 0.91 for video. The higher loss with video input is likely due to the fact that we are limited to using only 8 key frames (restricted by GPU memory; same as MotionLLM~\cite{chen2024motionllm} and Video-LLaVa~\cite{lin2023video}). Therefore, in general, for long sequences, high-quality motion is preferred.
For short videos without high-quality motion estimation, 
using video as input is more reliable due to information processing inequality.

\subsection{Ablation study on loss terms}.
We conducted additional experiments for $\mathcal{L}^{act}$ and $\mathcal{L}^{di s}$ separately as shown in Table~\ref{tab:reb}. 
Since the computation of $\mathcal{L}^{dis}$ depends on $\mathcal{L}^{act}$, we adjusted its calculation in the absence of $\mathcal{L}^{act}$ to directly take the input state as the input.
The experimental results show that removing either loss impacts performance. However, the absence of $\mathcal{L}^{dis}$ significantly hinders the network's ability to learn high-frequency information, leading to much less gain compared to removing $\mathcal{L}^{act}$.

\begin{table*}[t]
\caption{\small Additional ablation on specific loss terms.}
\vspace{-3mm}
\label{tab:reb}
\centering
\footnotesize
\begin{tabular}{@{}ccccc@{}}
\hline
\multirow{2}{*}{Model}      & \multirow{2}{*}{Overall} & \multicolumn{3}{c}{Query type}                   \\ \cline{3-5} 
                            &                          & Action         & Direction      & BodyPart       \\ \hline
Ours - w/o $\mathcal{L}^{act}$ & 0.563                    & 0.605          & 0.478          & 0.465          \\
Ours - w/o $\mathcal{L}^{dis}$ & 0.629                    & 0.683          & 0.457          & 0.577          \\ \hline
Ours                        & \textbf{0.711}           & \textbf{0.809} & \textbf{0.697} & \textbf{0.623} \\ \hline
\end{tabular}
\vspace{-6mm}
\end{table*}

\end{document}